\PassOptionsToPackage{hyphens}{url}
\documentclass[twocolumn]{reinpaper}
\usepackage{makecell}
\newcommand{\tool}{RL\textsuperscript{RT}}
\newcommand{\kbound}{\ensuremath{k}}
\DeclareRobustCommand{\supfull}{\tikz[baseline=-0.6ex]\fill (0,0) circle (0.7ex);}
\DeclareRobustCommand{\suppart}{\tikz[baseline=-0.6ex]{\draw[line width=0.4pt] (0,0) circle (0.7ex);\fill (0,-0.7ex) arc (-90:90:0.7ex) -- cycle;}}
\DeclareRobustCommand{\supnone}{\tikz[baseline=-0.6ex]\draw[line width=0.4pt] (0,0) circle (0.7ex);}
\newcommand{\takeaway}[2]{
\par\vspace{2.5mm}\noindent
\begin{tikzpicture}
\node[
 draw=black, line width=0.9pt, fill=white,
 rounded corners=1.2pt,
 inner xsep=6pt, inner ysep=6pt,
 text width=\dimexpr0.97\columnwidth-12pt\relax,
 align=left, font=\small
] (tkbody) {\rule[-0.3ex]{0pt}{3.4ex}\ignorespaces #2};
\node[
 fill=black, text=white, font=\footnotesize\bfseries,
 inner xsep=6pt, inner ysep=2.5pt,
 anchor=south west
] (tkhead) at ([xshift=4pt,yshift=0.3pt]tkbody.north west) {\strut #1};
\end{tikzpicture}\par\vspace{1.5mm}
}

\renewcommand{\reinlabname}{Nanyang Technological University}
\renewcommand{\reinlogo}[1][40pt]{\includegraphics[height=#1,keepaspectratio]{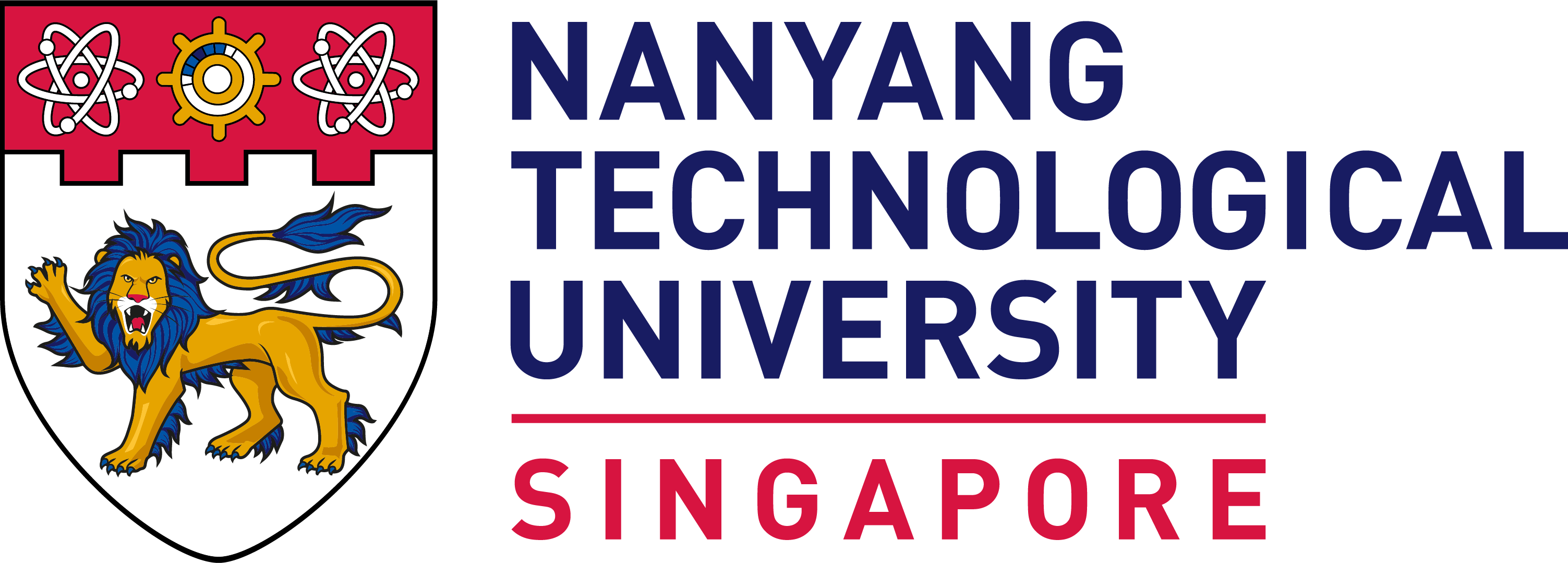}}
\renewcommand{\reinwordmark}{\reinlogo[40pt]}
\reinhidepartner
\makeatletter
\renewcommand{\rein@brandrow}{%
  \par\vspace*{20pt}\noindent\hspace*{20pt}\reinwordmark\par\vspace{20pt}}
\renewcommand{\rein@headerleft}{NTU SINGAPORE}
\renewcommand{\rein@footerleft}{RESEARCH PREPRINT}
\makeatother
\title{\tool: On-Device Reinforcement Learning Runtime}
\author{Zexin Li\reinorcid{0000-0001-8758-2151}\quad Xinwei Gao\quad Arvind Easwaran\reinorcid{0000-0002-9628-3847}}
\reinaffiliation{Nanyang Technological University, Singapore}
\reinemails{zexin.li@ntu.edu.sg; GAOX0039@e.ntu.edu.sg; arvinde@ntu.edu.sg}
\reinshorttitle{\tool: On-Device Reinforcement Learning Runtime}
\reintrack{EMBEDDED GPU SYSTEMS / REAL-TIME REINFORCEMENT LEARNING}
\reinstatus{RESEARCH PREPRINT}
\reinversion{}
\reinkeywords{on-device reinforcement learning; real-time scheduling; embedded GPUs; policy staleness}
\reingithub[github.com/ZexinLi0w0/rlrt]{https://github.com/ZexinLi0w0/rlrt}

\begin{document}
\maketitle
\begin{abstract}
On-device reinforcement learning runs a periodic actor and a learner on one embedded GPU shared with other workloads. Because the learner produces the actor's weights, existing designs stall the actor or let its policy drift. We present \tool, an on-device runtime that makes this coupling a tunable parameter: a refresh bound limits actor releases between refreshes, and a GPU permit with slack-aware admission prioritises the actor and fills spare time with training atop any base scheduler. Under the real refresh cost on two Jetson boards, \tool{} keeps a $16.8$\,M-parameter policy's deadline misses below $1\%$, against $85$ to $100\%$ without it.
\end{abstract}

\vspace{-12pt} 
\section{Introduction}

\begin{figure*}[!t]
\centering
\begin{minipage}[t]{0.355\textwidth}
\centering
\includegraphics[width=\linewidth]{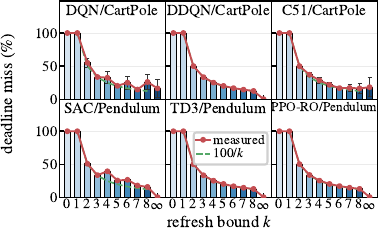}
\caption{Actor miss rate on AGX against refresh interval $k$ processed actor releases; dashed: $(100/k)\%$ reference for $k\ge1$.}
\label{fig:e4miss}
\end{minipage}\hfill
\begin{minipage}[t]{0.355\textwidth}
\centering
\includegraphics[width=\linewidth]{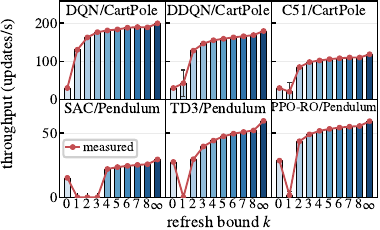}
\caption{Learner throughput in the same runs; $k=0$ is lockstep and $k=\infty$ is free-running.}
\label{fig:e4qps}
\end{minipage}\hfill
\begin{minipage}[t]{0.27\textwidth}
\centering
\includegraphics[width=\linewidth]{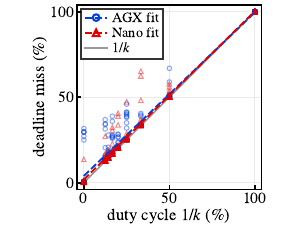}
\caption{Miss rate against refresh duty cycle in $600$ runs.}
\label{fig:e13}
\end{minipage}
\end{figure*}

As deep reinforcement learning moves to robots and vehicles
that act in the physical world~\cite{mnih2015human,d3}
and must keep learning while operational, it needs a runtime that coordinates inference, training and policy updates on
one embedded accelerator. A periodic actor must deliver actions by its control deadline, a learner needs enough compute time to update the policy,
and both share the GPU with other
workloads~\cite{kang2021lalarand,li2021efficient}.

The actor and learner compete for compute while depending on each
other, so the runtime must coordinate policy freshness with actor timeliness and
learner throughput. Learning benefits control only when the learner's weights
reach the actor~\cite{horgan2018distributed,espeholt2018impala}, yet
waiting for training or a policy refresh can delay an action beyond its
deadline~\cite{ramstedt2019real}, and an independent actor runs on increasingly stale weights.
GPU isolation and scheduling limit interference between co-located
tasks~\cite{elliott2013gpusync,s3dnn} but do not specify when
the actor should receive new weights.

Our fundamental research challenge is the following.
\vfill
\takeaway{Fundamental research challenge}{How often should a firm-deadline actor refresh its policy, and when should the learner run on the shared embedded GPU, so that learning keeps the policy fresh while the actor keeps deadline misses low?}
\newpage

Current designs fix the coupling at one end or leave it implicit, without
stating what is being traded.
Lockstep designs synchronise
the actor with every published update, keeping the policy fresh at a synchronisation cost on every release. In our
measurements a lockstep learner is throttled to the actor's rate, up to $6.7\times$ (Section~\ref{sec:eval-real}) below what the device sustains, and once the refresh or the learner step no longer fits in the actor's slack the actor misses essentially every deadline. At the other end, an actor that never waits for the learner protects its timing but leaves staleness unbounded; free-running never refreshes and drifts thousands of updates behind. Asynchronous learners such as A3C~\cite{mnih2016asynchronous}, Ape-X~\cite{horgan2018distributed} and IMPALA~\cite{espeholt2018impala} sit between the two by refreshing periodically, but leave the interval as an implicit tuning choice.

We present \tool, an on-device reinforcement learning runtime that makes
this trade an explicit, tunable and measurable runtime parameter. A refresh gate bounds how many actor releases may
pass between refresh attempts, placing lockstep and free-running at opposite
ends of one axis. A GPU permit serialises work from the actor, learner and
third-party tenant, and a slack-aware admission layer, atop the platform's
real-time scheduler, gives the actor priority and fills its spare time with
learner steps, splitting updates too long to fit. Because the base ordering is a parameter, \tool{} is a runtime orthogonal to GPU and on-device RL schedulers such as GCAPS~\cite{gcaps_ecrts24} and REEF~\cite{han2022reef} and runs their request orderings as its base policy. Under the real refresh cost, the permit and admission set the actor's timing and the bound mainly sets update lag.

We evaluate \tool{} on an Orin AGX and an Orin Nano
across six reinforcement learning algorithms, two control environments, six base schedulers, a swept actor deadline and
actor networks from $17.4$\,k to $16.8$\,M parameters, and measure the runtime's own cost
separately. With the real
refresh cost on a $16.8$\,M-parameter replay-based actor-critic policy, it kept measured misses below
$1\%$ where a runtime-free actor misses $85$ to $100\%$, at lower learner
throughput. As a feasibility demonstration, we also integrate \tool{} into the full visual pipeline of a stationary Duckiebot~\cite{paull2017duckietown}, where on-device imitation learning shares the GPU with a $100$\,ms control loop, and we will
release the implemented library.

This paper makes the following contributions:
\begin{itemize}[leftmargin=10pt]
\item We provide a runtime for on-device reinforcement learning that makes the actor--learner coupling an explicit runtime parameter, a bound on the releases between refresh attempts, and co-schedules a firm periodic actor, an elastic learner and a third-party tenant on one embedded GPU with a permit, slack-aware admission and update splitting atop existing schedulers.
\item We give a calibration procedure that derives a feasible actor deadline
      and the refresh bound from measured reference runs, and measure the real refresh cost.
\item We comprehensively evaluate the runtime on two Jetson platforms along the algorithm,
      scheduler, deadline, and model-size axes, including Atari
      Breakout, and integrate it into the visual pipeline of a stationary
      robot.
\end{itemize}

\section{Background and Motivation}
\label{sec:bg}

\subsection{On-device Reinforcement Learning}

On-device deep RL shares one accelerator between a periodic \emph{actor}
with a firm deadline $D$ no later than its period $T$ and a throughput-oriented \emph{learner}
that updates the policy from experience. Off-policy methods for discrete and continuous control~\cite{mnih2015human,van2016deep,bellemare2017distributional,haarnoja2018soft,fujimoto2018addressing}
learn from older policies' data and need many gradient updates; on robots the obstacle is rarely the algorithm~\cite{gu2017deep}
but sharing the GPU, which creates both execution interference and a dependency:
each refresh of the actor's weights from the learner can add synchronisation and transfer costs to the actor's release.

\subsection{Metrics}
\label{sec:metrics}

We measure actor timeliness, learner throughput and policy staleness on the board during each co-run, evaluating the deployed system dynamically rather than from a static profile~\cite{chen2025benchmarking}.
For $N$ actor releases at times $r_i$, with execution times $e_i$ and
completion times $f_i$ over a run of length $L$, let $R_i=f_i-r_i$.
{Deadline-miss rate} is $M=N^{-1}\bigl|\{\,i:R_i>D\,\}\bigr|$.
{Tail step time} is the empirical execution-time quantile
\begin{equation}
Q_q=\inf\{x:N^{-1}|\{i:e_i\le x\}|\ge q\},
\label{eq:m}
\end{equation}
for
$q\in\{0.9,0.99,0.999\}$; it excludes GPU waiting time $R_i-e_i$.
Learner throughput is $\Lambda=v_\ell(L)/L$, where
$v_\ell(t)$ counts completed learner updates, each advancing the
policy version by one from zero. In particular, subscripts $\ell$ and $a$ denote
learner and actor; $v_a(t)$ is the version currently held by the actor.
{Staleness} has two measures: adoption age $n_i$, the number of
processed actor releases since the last successful weight adoption,
and update lag $S_i=v_\ell(t_i)-v_a(t_i)$, the number of learner
updates by which the actor's policy trails the learner's.
Here $t_i$ is sampled after any refresh attempt and before
requesting the GPU permit for actor step $i$.

\subsection{Bounding Staleness Instead of Interference}

Rather than suppress learner interference, which prevents learning, or
refresh at every release, which puts synchronisation on every deadline's
critical path, \tool{} exposes the refresh interval $\kbound$ as a timing
parameter: after $\kbound$ processed releases since its last adoption, the actor tries to adopt the newest complete snapshot. Lockstep ($\kbound=0$) also waits for a fresh
update, free-running ($\kbound=\infty$) keeps the initial weights, and update lag scales roughly as $\kbound T\Lambda$. Asynchronous RL could already refresh
periodically~\cite{mnih2016asynchronous,horgan2018distributed}.

\noindent\textbf{Motivational study.}
As a controlled study of the gate, with a refresh cost injected to miss the deadline, the actor's miss rate
tracks the refresh duty cycle $1/\kbound$; Section~\ref{sec:eval-real} measures the real cost.
Figures~\ref{fig:e4miss} and~\ref{fig:e4qps} sweep $\kbound$ for six
algorithms on the AGX with three tenants and the injected cost of
Section~\ref{sec:design}, and learner throughput generally rises from lockstep. Over $600$ runs on both
boards the miss rate regresses on the duty cycle with slope $0.97$
and coefficient of determination $R^2=0.98$ (Figure~\ref{fig:e13}), confirming that the gate refreshes at the configured rate; tight intervals can starve the
learner, and a phase-locked third-party tenant causes misses without
refreshes. One miss per refresh approximates weakly-hard behaviour:
at most one miss per $\kbound$ consecutive releases~\cite{choi2021toward}.

\subsection{Relation to prior work}

Table~\ref{tab:related} contrasts \tool{} with the closest systems. GPU real-time systems make sharing predictable through
reservations~\cite{elliott2013gpusync},
kernel scheduling~\cite{s3dnn}, CPU--GPU
partitioning~\cite{bib:RTSS19:Xiang,kang2021lalarand} and
preemption~\cite{han2022reef,gcaps_ecrts24}, and other systems trade inference quality
for timeliness~\cite{d3,bateni2018apnet}; \tool{} instead trades
how often the actor refreshes. R$^3$~\cite{li2023r3} adapts batch and
replay sizes to device budgets through deadline feedback, sizing each training step, whereas the base policies of Figure~\ref{fig:e9} order GPU requests, and real-time RL~\cite{ramstedt2019real}
models action delay; neither arbitrates the GPU between actor and
learner. Distributed RL~\cite{mnih2016asynchronous,espeholt2018impala,horgan2018distributed}
and stale synchronous training~\cite{ho2013more} tolerate stale parameters
for throughput, and RLlib and Sample Factory (SF), the frameworks measured in Section~\ref{sec:eval-baselines}, maximise throughput without an actor deadline.

\section{\tool{}: A Deadline-Aware Runtime for Co-Located RL Training and Inference}
\label{sec:design}

\tool{} coordinates the actor, learner and third-party tenant through a shared
GPU permit, a learner admission rule and a periodic refresh gate. Each thread submits a step request with an estimated cost and, for periodic
tenants, release and deadline information.

\noindent\textbf{Objective.} Given a deadline-miss budget $M_{\max}$ for the actor, \tool{} keeps the policy as fresh as the budget allows. With $\bar{S}$ the mean update lag of Section~\ref{sec:metrics}, it solves
\begin{equation}
\min_{\kbound}\ \bar{S}(\kbound)\quad \text{s.t.}\quad M(\kbound)\le M_{\max},
\label{eq:obj}
\end{equation}
and, for the chosen $\kbound$, admission maximises the learner throughput $\Lambda$ over learner steps that fit the actor's slack. Since lag grows with $\kbound$, calibration (Section~\ref{sec:calib}) selects the smallest $\kbound$ meeting the budget.

\begin{table}[t]
\centering
\caption{Comparison with existing methods. \supfull, \suppart, and \supnone{} denote support, partial support, and no support; N/A: not applicable. Actor deadline: takes a deadline and lowers measured misses.}
\label{tab:related}
\footnotesize
\setlength{\tabcolsep}{2.0pt}
\begin{tabular}{@{}lcccccc@{}}
\toprule
\textbf{Method} & \textbf{Target} & \textbf{\makecell{Actor\\deadline}} & \textbf{\makecell{Refresh\\bound}} & \textbf{\makecell{GPU\\arbiter}} & \textbf{\makecell{Slack\\admission}} & \textbf{\makecell{Update\\split}} \\
\midrule
R$^3$~\cite{li2023r3} & Embedded & \supfull & \supnone & \supnone & \suppart & \suppart \\
GPUSync~\cite{elliott2013gpusync} & Multi-GPU & \supfull & N/A & \supfull & N/A & N/A \\
RLlib~\cite{liang2018rllib} & Cluster & \supnone & \suppart & \supnone & \supnone & \supnone \\
SF~\cite{petrenko2020sample} & Server & \supnone & \suppart & \supnone & \supnone & \supnone \\
\midrule
\tool{} & Embedded & \supfull & \supfull & \supfull & \supfull & \supfull \\
\bottomrule
\end{tabular}
\end{table}

\begin{algorithm}[t]
\footnotesize
\caption{\tool{} runtime}
\label{alg:rlrt}
\begin{algorithmic}[1]
\Require interval $\kbound$, margin $\mu$, lent slack $\lambda$, scheduler $\sigma$
\Require current time $t$, predicted cost $\hat{c}(r)$ of request $r$
\Require learner weights $W$, actor weights $\theta$ (policy $\pi_\theta$), replay buffer $\mathcal{D}$, periodic-tenant registry $\mathcal{P}$
\Require $\mathit{next}(p,t)$: tenant $p$'s next release at or after $t$
\State $v_\ell \gets 0$; $v_a \gets 0$; $n \gets 0$; $\theta \gets W$ \Comment{$n$: adoption age}
\State \textbf{proc} \textsc{Refresh} \Comment{$\mathit{ok}$: adoption succeeded}
\State \quad $(\theta,v_a,\mathit{ok}) \gets \textsc{AdoptLatest}(\theta,v_a)$; \textbf{if} $\mathit{ok}$ \textbf{then} $n \gets 0$
\smallskip
\State \textbf{proc} \textsc{Actor}$(s)$ \Comment{observation $s$ at each processed release}
\State \quad $n \gets n+1$
\State \quad \textbf{if} $\kbound=0$ \textbf{then}
        \textbf{await} $v_\ell>v_a$; \textsc{Refresh}
\State \quad \textbf{else if} $\kbound<\infty \wedge n\ge\kbound$
        \textbf{then} \textsc{Refresh}
\State \quad \textsc{Acquire}; action $a \gets \pi_\theta(s)$; \textsc{Sync}; \textsc{Release}
\smallskip
\State \textbf{proc} \textsc{Learner} \quad \Comment{each iteration}
\State \quad \textbf{if} $\kbound=0$ \textbf{then await} $v_a=v_\ell$
\State \quad minibatch $B\sim\mathcal{D}$; $(u_1,\ldots,u_m)\gets\textsc{SplitUpdate}(B)$ \Comment{$m$ pieces}
\State \quad \textbf{for} $j=1,\ldots,m$ \textbf{do}
\State \qquad \textsc{Acquire}$(u_j)$; $W\gets u_j(W)$; \textsc{Sync}
\State \qquad \textbf{if} $j=m$ \textbf{then} \textsc{Publish}$(W,v_\ell+1)$
\State \qquad \textsc{Release}
\smallskip
\State \textbf{proc} \textsc{Grant}$(\mathcal{R},t)$ \Comment{pending requests $\mathcal{R}$; permit free}
\State \quad $\mathcal{R}_{\mathrm{rt}}\gets\{r\in\mathcal{R}:r.\mathit{kind}\ne\ell\}$ \Comment{kind: tenant type}
\State \quad \textbf{if} $\mathcal{R}_{\mathrm{rt}}\ne\emptyset$ \textbf{then return} $\sigma(\mathcal{R}_{\mathrm{rt}})$
\State \quad \textbf{if} $\mathcal{R}=\emptyset$ \textbf{then return} $\bot$ \Comment{no grant}
\State \quad $r_\ell\gets$ the learner request in $\mathcal{R}$
\State \quad \textbf{if} $\kbound=0\wedge$ actor awaits $r_\ell$
        \textbf{then return} $r_\ell$
\State \quad $h\gets\min_{p\in\mathcal{P}}\mathit{next}(p,t)$ \Comment{$h=\infty$ if $\mathcal{P}=\emptyset$}
\State \quad \textbf{if} $t+\hat{c}(r_\ell)+\mu\le h+\lambda$
        \textbf{then return} $r_\ell$
\State \quad \textbf{return} $\bot$ \quad \Comment{defer}
\end{algorithmic}
\end{algorithm}

\subsection{Mechanisms}

\noindent\textbf{Permit.} The permit serialises GPU work so that kernel
interleaving cannot stretch actor execution, trading concurrency for
predictability. The runtime grants one step at a time, and each step
synchronises its CUDA stream before releasing the permit. The permit coordinates every tenant that requests it. Although Orin supports context preemption, stock JetPack
exposes no user-level priority preemption, and CUDA stream priorities only
order pending work~\cite{han2022reef,gcaps_ecrts24}. The permit therefore
controls when steps start, using established GPU
locking~\cite{elliott2013gpusync}.%

\noindent\textbf{Admission rule.} Ready real-time requests take priority in
base-policy order; otherwise a learner request $r_\ell$ starts only if
\begin{equation}
 t + \hat{c}(r_\ell) + \mu \le h + \lambda,
\label{eq:admit}
\end{equation}
where $t$ is current time, $\hat{c}(r)$ predicts request $r$'s cost, $h$ is
the next periodic release, $\mu$ a safety margin and $\lambda$ slack lent
past $h$. After five observations, costs use the median of the last $2048$
step times. An awaited lockstep update bypasses admission to avoid deadlock.
Long updates split into minibatch gradient-accumulation passes and optimiser
steps over parameter groups. Each piece is admitted separately using the
median cost of recent pieces; only complete updates are published.
Equivalence to unsplit updates depends on the adapter
(Section~\ref{sec:eval-baselines}). Admission is empirical, not a guarantee:
in the sweep of Section~\ref{sec:bg}, learner steps stay within
$\hat{c}(r_\ell)+\mu$ in $95.7$ and $92.4\%$ of cases on the AGX and the
Nano, and $99\%$ finish within $12$\,ms of it.

\noindent\textbf{Refresh gate.} The gate bounds how many releases pass
between refresh attempts, and each refresh takes one of two paths. At $\kbound=0$, actor and learner alternate:
the actor waits for a published update, and the learner waits for it to be
consumed before updating again. For finite $\kbound\ge1$, the actor attempts a refresh
after $\kbound$ processed releases since its last adoption, retrying at each later release, without blocking the learner; at $\kbound=\infty$,
it keeps its initial weights. With double buffering, the learner
copies completed weights into a shadow buffer under a lock within its permit
slot; the actor swaps in that whole version at a release, retaining its
weights and retrying at the next release if no newer version exists. Publication and swap share one non-blocking lock, so the learner overwrites only a buffer the actor has released; a learner that finds it held skips that copy until its next update, and an actor that finds it held keeps its weights and retries at its next release. The published version counts completed updates, and the next completed update refreshes the shadow buffer. Split updates use this path. On the copy
path, the actor copies weights with a device barrier before requesting its
permit. This cost is charged to its response, and the copy can overlap a
running learner step, whereas double buffering adopts only complete snapshots off the actor's critical path, so $\kbound\ge1$ mainly controls
update lag.

\noindent\textbf{Calibration and interval selection.}\label{sec:calib} Measured reference
runs set $D$, the injected refresh cost and the deployed interval. For each board, model
and workload, reference co-runs with $\kbound=\infty$ set $D$ above the
actor's measured 99th-percentile response, leaving room within the period
for a refresh. The controlled sweeps additionally inject GPU work of cost
$c_r$, sized from the median response to push refreshing releases past $D$
but inside the period, a controlled stand-in for the copy cost that Section~\ref{sec:eval-real} measures. Since
admission depends on $D$, calibration repeats at the derived deadline until
it stabilises or reaches the iteration limit, flagging configurations whose
tail does not fit. On the deployed refresh path, the runtime selects $\kbound=1$ if the
refreshing response's 99th percentile meets $D$. Otherwise, assuming each
refresh causes one miss, it chooses the smallest interval meeting the
control loop's miss budget $M_{\max}$,
$\kbound=\lceil 1/M_{\max} \rceil$. If refreshes cause the only misses, this empirical selection predicts an average miss fraction near $1/\kbound$, which the evaluation measures.

\subsection{Algorithm}

Algorithm~\ref{alg:rlrt} combines the mechanisms; updates use $m$ pieces
($m=1$ without splitting).
\textsc{Grant} runs on request arrivals and step completions.
With $N_{\mathrm t}$ tenants, scheduling takes $O(N_{\mathrm t}\log N_{\mathrm t})$ and the gate
$O(1)$ time, excluding GPU work.

\begin{table}[!tbp]
\centering
\caption{Learner updates/s / actor miss (\%). Small/large: $34.3$\,k/$16.8$\,M parameters. Bold/underline: best/second best excluding the grey baseline. \tool{} uses double buffering; baseline uses copy.}
\label{tab:e16}
\footnotesize
\setlength{\tabcolsep}{2.2pt}
\begin{tabular}{lcccc}
\toprule
 & \multicolumn{2}{c}{Orin Nano} & \multicolumn{2}{c}{Orin AGX} \\
\cmidrule(lr){2-3}\cmidrule(lr){4-5}
 & small & large & small & large \\
\midrule
\textcolor{gray}{no runtime} & \textcolor{gray}{71.6 / 0.6} & \textcolor{gray}{11.8 / 85.3} & \textcolor{gray}{90.0 / 0.4} & \textcolor{gray}{30.0 / 99.9} \\
\midrule
RLlib~\cite{liang2018rllib} & 8.1 / \underline{3.0} & \textbf{5.5} / \underline{1.6} & \underline{11.9} / \underline{1.1} & \underline{9.8} / \underline{3.1} \\
SF~\cite{petrenko2020sample} & \underline{9.2} / 16.1 & 3.1 / 22.5 & 9.2 / 10.9 & 3.4 / 11.1 \\
\midrule
\tool{} & \textbf{55.1} / \textbf{0.7} & \underline{4.7} / \textbf{0.6} & \textbf{60.0} / \textbf{0.5} & \textbf{23.7} / \textbf{0.5} \\
\bottomrule
\end{tabular}
\end{table}

\subsection{System Prototype}

\tool{} comprises about $4$\,k lines of Python on PyTorch, packaged as
\texttt{librlrt}. Adapters for DQN, DDQN, C51, SAC, TD3 and PPO-RO (Section~\ref{sec:eval}) share replay and weight publication.
Existing training loops adopt the runtime by wrapping actor and learner
steps with its step-request API. Adapters supply these steps, the published weights and any update split. Per-request timestamps
separate permit, gate and telemetry overhead (Section~\ref{sec:eval-overhead}).
We will release \texttt{librlrt} upon acceptance, together with a compiled C/C++ core of the permit, admission rule and gate in a shared library with the same Python API on top.

\setcounter{topnumber}{2}
\begin{figure}[t]
\centering
\includegraphics[width=\columnwidth]{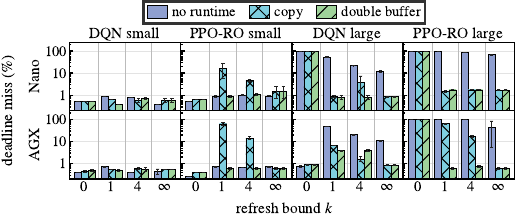}
\caption{Actor deadline miss with the real refresh cost, mean
and standard deviation (SD) of three seeds. Small: $17.4$\,k (DQN) and $34.3$\,k (PPO-RO) parameters;
large: $16.8$\,M.}
\label{fig:e14}
\end{figure}

\begin{figure}[t]
\centering
\includegraphics[width=0.44\textwidth]{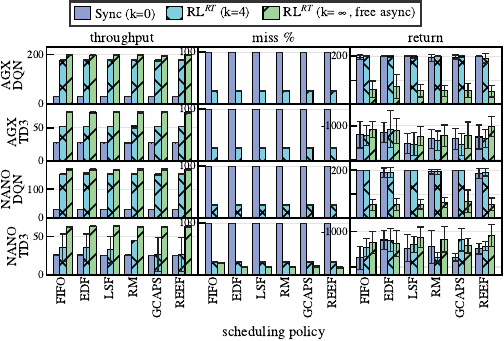}
\caption{Effectiveness under six base schedulers on two Jetson boards. LSF: least-slack-first; RM: rate-monotonic.}
\label{fig:e9}
\end{figure}

\begin{figure}[t]
\centering
\includegraphics[width=0.45\textwidth]{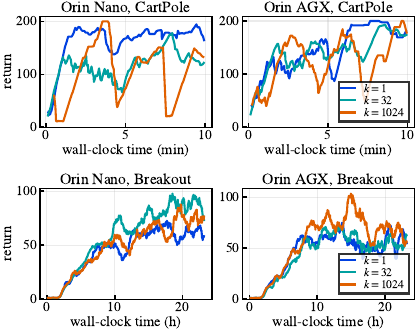}
\caption{Return against wall-clock time on the copy refresh path.}
\label{fig:e15}
\end{figure}

\begin{figure}[t]
\centering
\includegraphics[width=0.45\textwidth]{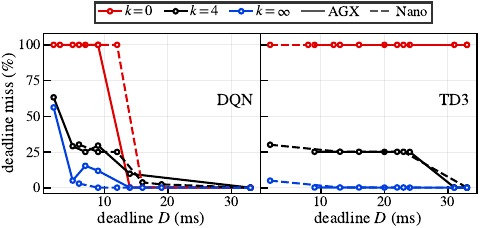}
\caption{Actor miss rate against the deadline the actor is given, with the
refresh cost held at its calibrated value.}
\label{fig:e10}
\end{figure}

\begin{figure}[t]
\centering
\begin{minipage}[t]{0.49\columnwidth}
\centering
\includegraphics[width=\linewidth,height=1.93in,keepaspectratio]{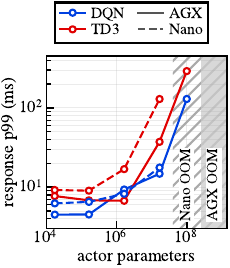}
\caption{Contended response $p99$ (99th percentile) against actor size; OOM: out of memory.}
\label{fig:e11}
\end{minipage}\hfill
\begin{minipage}[t]{0.49\columnwidth}
\centering
\includegraphics[width=\linewidth,height=1.93in,keepaspectratio]{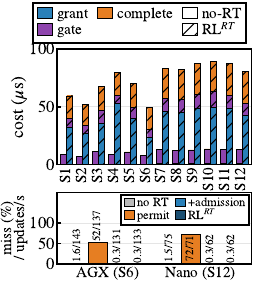}
\caption{Step cost (top), ablation (bottom). S1--S6 (AGX), S7--S12 (Nano): DQN, DDQN, C51, SAC, TD3, $1024$-wide MLP. Grant/gate/complete: acquisition/refresh/bookkeeping.}
\label{fig:e8}
\end{minipage}
\end{figure}

\begin{figure}[t]
\centering
\includegraphics[width=0.45\textwidth]{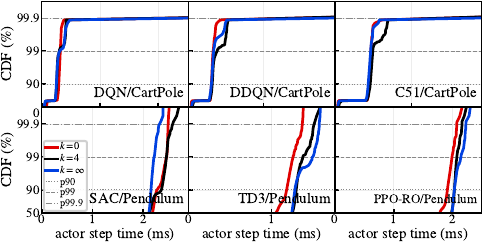}
\caption{Actor execution-time cumulative distribution function (CDF) on Orin AGX; $p90$, $p99$, $p99.9$ mark the 90th, 99th, 99.9th percentiles.}
\label{fig:e12}
\end{figure}

\newpage
\section{Evaluation}
\label{sec:eval}

\subsection{Setup}

\noindent\textbf{Hardware and software.} All experiments except the
real-world case study (Section~\ref{sec:casestudy}), which runs on the robot's
own Jetson Nano 2GB, use two Jetson platforms, an Orin AGX with $32$\,GB and its GPU pinned at
$1300.5$\,MHz, and an Orin Nano with $8$\,GB pinned at $612$\,MHz. Clocks are verified before every
run, and no thermal throttling occurred in these experiments. Both boards run JetPack~6.2 (CUDA~12.6) with PyTorch~2.8 and
Python~3.10.

\noindent\textbf{Workloads.} Unless stated otherwise, all
experiments share the following settings. The actor
runs at $30$\,Hz with an update-to-data cap of $8$. Workloads include
DQN~\cite{mnih2015human}, DDQN~\cite{van2016deep},
C51~\cite{bellemare2017distributional}, SAC~\cite{haarnoja2018soft},
TD3~\cite{fujimoto2018addressing} and a replay-based PPO
variant~\cite{schulman2017proximal} that keeps PPO's networks,
optimiser and loss coefficients but, lacking behaviour log-probabilities, has ratio one and so yields asynchronous advantage actor-critic updates; we call it PPO-RO. A third tenant runs matrix workloads at $30$\,Hz
with a period-length deadline: $1152^2$ on AGX and $384^2$ on Nano,
taking $8.6$ and $2.1$\,ms, or $26$ and $6\%$ of the GPU.

\noindent\textbf{Baselines.} Baselines are lockstep, which runs one update and refresh per release, and free-running, which never blocks the actor or refreshes. Section~\ref{sec:eval-baselines} also compares RLlib's
separate Ray actor/learner processes~\cite{liang2018rllib} and Sample Factory's
asynchronous PPO trainer~\cite{petrenko2020sample}. R$^3$~\cite{li2023r3} sizes training steps, which is orthogonal to GPU arbitration, and the permit-alone arm of Figure~\ref{fig:e8} is a locking abstraction like GPUSync's~\cite{elliott2013gpusync}.
Late actions are applied; releases skipped while the actor is busy count
as misses. Runs last $25$\,s with three to five seeds by default.

\noindent\textbf{Runtime parameters.} Admission uses $\mu=0.5$\,ms and empirically chosen $\lambda=3$\,ms (AGX)
or $6$\,ms (Nano). Calibration sets $D$ to the rounded-up $1.25\times$
reference 99th percentile, capped at $T-8$\,ms. The injected $c_r$ is
$D$ minus the median response plus $5$\,ms, clamped to $[4\,\mathrm{ms}, T-\mathrm{median}-3\,\mathrm{ms}]$, and calibration
runs at most three iterations of three $25$\,s references. The robot of Section~\ref{sec:casestudy} uses its $100$\,ms application deadline.

\subsection{Overall Effectiveness}

\noindent\textbf{Comparison with existing frameworks.}\label{sec:eval-baselines}
Against RLlib and SF, \tool{} has the lowest actor miss rate in every
configuration of Table~\ref{tab:e16}, under $1\%$ throughout, and the highest
learner throughput in three of four. The table runs \tool{} at $\kbound=4$ on its replay-based PPO-RO and RLlib and SF on their native APPO with $34.3$\,k and $16.8$\,M parameters, each framework at its highest-throughput setting of batch reuse $1000$ for RLlib and $100$ epochs per batch for SF, as the mean of three seeds. In every system an update is one Adam step on $256$ samples, so the table compares runtime configurations at a matched update size; RLlib~2.59 and SF~2.1.1 run APPO with one environment and a $32$-step rollout on the GPU. The runtime-free asynchronous actor, the grey row at $\kbound=4$ on the copy path, reaches the highest throughput but
misses $85$ to $100\%$ of deadlines on the large network, while on the small one it matches \tool{}'s misses at more throughput, so the runtime pays off once learner steps contend with the actor. Serialising
through the permit costs $21$ to $33\%$ of the grey row's learner updates/s
when updates run whole and $60\%$ when they are split {, the price of $0.5\%$ instead of $85$ to $100\%$ misses on the large network}. On the Nano's large network, \tool{} trains by splitting each $80$\,ms update into admitted pieces: eight $32$-sample passes of $16$\,ms accumulate the gradient, and three optimiser pieces apply Adam to parameter groups. With one pass, a split update matches an unsplit one to $1.5\times10^{-8}$, and with eight it differs only in normalising advantages per piece; without splitting, \tool{} there trains at $0.0$ updates/s. The third-party tenant misses at most $0.7\%$ of its deadlines under \tool{}, against $18.8\%$ without it on the Nano's large network.

\noindent\textbf{Real refresh cost.}\label{sec:eval-real}
Under the real refresh cost, a runtime-free actor matches \tool{} on small
networks but misses far more at $16.8$\,M parameters. Figure~\ref{fig:e14}
repeats the sweep of Section~\ref{sec:bg} with the copy-and-barrier cost over $351$ runs per board, adding double buffering and a
runtime-free actor on prioritised CUDA streams. A Nano copy costs under
$0.7$\,ms up to $1.65$\,M parameters and $10.9$\,ms at $67$\,M, so the injected $c_r$ matches about $65$\,M; the AGX tenant is
offset by half a period against $30$\,Hz phase-locking. On small networks, only PPO-RO's copy path misses many deadlines under \tool{}, $16.3$ and $63.1\%$ at $\kbound=1$, which both baselines avoid, and lockstep caps throughput at $30$ against up to $202$ updates/s. At
$16.8$\,M parameters the runtime-free actor misses $11$ to $100\%$, against $1$ to $7\%$ for DQN
under \tool{} and under $1\%$ for double-buffered PPO-RO on the AGX. With double buffering, $\kbound$ mainly sets the update lag: at $\kbound \ge 1$ the miss rate stays within $0.5$ points of free-running in $17$ of $18$ configurations, the exception being DQN at $16.8$\,M on the AGX at $3.8$ to $4.0$ against $0.8\%$.

\noindent\textbf{Composition with base schedulers.}\label{sec:eval-sched}
The admission layer composes with the platform's existing ordering, and the trade is the same under six base orderings of permit requests over $180$ runs per board with the injected refresh cost in Figure~\ref{fig:e9}: FIFO, EDF, least-slack-first, rate-monotonic, GCAPS~\cite{gcaps_ecrts24} and REEF~\cite{han2022reef}. GCAPS orders by criticality, then relative deadline, then arrival, and REEF serves requests with a deadline before best-effort ones, each class in arrival order. All six are user-space orderings on stock JetPack with run-to-completion grants, so the comparison isolates the ordering rule that each scheduler contributes. Under all six, lockstep is throughput-starved and misses almost everything, $\kbound=4$ recovers most of the throughput at the injected duty cycle of one refresh in four releases, and free-running maximises it; the miss rate at $\kbound=4$ varies by at most $0.06$ points across them. \tool{} is thus orthogonal to the base ordering.

\noindent\textbf{Learning across refresh intervals.}\label{sec:eval-lag}
Figure~\ref{fig:e15} trains DQN with the real refresh cost on the copy path on both boards at
$\kbound = 1$, $32$ and $1024$, showing seed $0$: CartPole for $10$\,min and
Atari Breakout for $10$\,M frames, about $23$\,h, with the DQN settings of
R$^3$~\cite{li2023r3}. On Breakout every bound learns, ending at a last-$100$-episode mean training return of $55$ to $76$ with no consistent order
across the boards, while the update lag reaches $256$ updates at
$\kbound=1024$. On CartPole the lag at $\kbound = 1024$ reaches about $6{,}000$ updates, and the last-$10$-episode mean return (mean $\pm$ SD over three seeds) falls to
$122 \pm 12$ on the Nano against $161 \pm 11$ at $\kbound = 1$ and to $137 \pm 37$
against $152 \pm 39$ on the AGX, whereas $\kbound = 32$ stays within the seed
spread. Refreshing at every release misses $1.1$ and $5.8\%$ of CartPole deadlines, against at most $0.4\%$ at larger bounds, and under $0.01\%$ on Breakout.

\subsection{Ablation Study}

\noindent\textbf{Deadline.} The duty-cycle behaviour holds over a window of deadlines: in the deadline sweep of Figure~\ref{fig:e10} with the refresh cost fixed, $325$ runs on
the AGX and $335$ on the Nano, Nano TD3 at $\kbound=4$
misses $25.2$ to $25.3\%$ from $D = 13$ to $23$\,ms, so a deadline several milliseconds off keeps the $1/\kbound$ pattern.

\noindent\textbf{Model size.} Up to $16.8$\,M parameters for discrete control and $1.65$\,M for continuous
control, the runtime's behaviour does not change with model size on either board in Figure~\ref{fig:e11}, with calibration re-run per size, and its
own cost stays at $55$ to $56\,\mu$s. Beyond it the contended tail reaches $130$ and $291$\,ms at
$10^8$ parameters on the AGX, so no $\kbound$ recovers the deadline; the
learner state no longer fits in memory from $10^8$ parameters on the Nano and
from $484$\,M on the AGX. The sweep injects the refresh cost and runs updates whole.

\noindent\textbf{Permit and admission.} Figure~\ref{fig:e8} (bottom) separates the mechanisms on a $1024$-wide learner without refreshes, S6 on the AGX and S12 on the Nano: no runtime misses $1.6$ and $1.5\%$ of deadlines, the permit alone $52$ and $72\%$, and admission $0.3\%$ at $131$ and $62$ learner updates/s against $137$ and $71$ for the permit alone. In the $16.8$\,M-parameter PPO-RO configuration of Table~\ref{tab:e16}, \tool{} misses $0.55$ to $0.63\%$ at $\kbound=1$, $4$ and $\infty$, while the runtime-free actor misses $49.5\%$ on the AGX and $70.4\%$ on the Nano at $\kbound=\infty$, where it never refreshes and the refresh path plays no part, and the $99.9$ and $85.3\%$ of Table~\ref{tab:e16} at $\kbound=4$, where it refreshes. Admission costs $5$ to $14\%$ more energy per update than no runtime, at $41$ to $43$\,W on the AGX and $11$ to $12$\,W on the Nano.

\subsection{Overhead Analysis}
\label{sec:eval-overhead}

\noindent\textbf{Per-step cost.} \tool{} adds $50$ to $89\,\mu$s per actor step, under $6\%$ of the median
response and under $4\%$ of its $p99$. Figure~\ref{fig:e8} measures it against a hand-written three-thread baseline without permit,
policy or telemetry for DQN, DDQN, C51, SAC, TD3 and a $1024$-wide learner on each board. In a CPU-only policy-decision microbenchmark on the Nano, timed inside each backend, the C/C++ core cuts a decision cost  from $3.4$ to $0.14\,\mu$s.

\noindent\textbf{Tail step time.} In Figure~\ref{fig:e12}, actors reach at most $5.6$\,ms at $p99.9$ and $13.4$\,ms at worst on both boards, against deadlines of at least $6$\,ms, so the deadline mainly absorbs the refresh.

\subsection{Practical Case Study}
\label{sec:casestudy}

\begin{figure}[!t]
\centering
\includegraphics[width=\columnwidth]{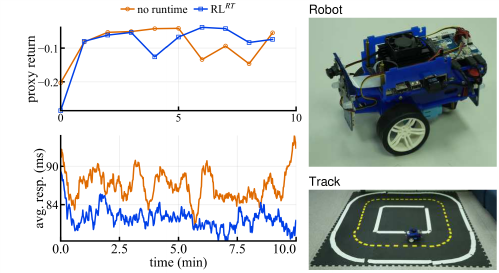}
\caption{Left: offline imitation proxy return (top) and actor response time
(bottom; 30\,s mean). Right: robot and track used in this study.}
\label{fig:casestudy-synth}
\end{figure}

\noindent\textbf{Robots Setup.} As a feasibility demonstration, we evaluate the raised, stationary Duckiebot testbench~\cite{paull2017duckietown} on the right of Figure~\ref{fig:casestudy-synth}, whose Jetson
Nano 2GB runs its complete ROS2 visual pipeline and a pretrained 2.24M-parameter
ACNet in CUDA FP16, while its Raspberry Pi runs the ROS 2 motor-control node. Both arms run the same batch-1 offline imitation on 1,024 recorded samples
with an actor MSE loss and an auxiliary Huber critic, Adam and gradient clipping; the learner has no deadline. The actor runs at $\kbound=4$. Wheel commands publish the latest action at 30\,Hz; each arm
runs 30\,s of warmup and 600\,s.

\noindent\textbf{Response time.} In a paired run, \tool{} lowers the actor's mean response by 5.87\% from 86.96 to
81.86\,ms and its misses of the 100\,ms deadline from 27.2 to 22.1\% of processed responses, with p99 falling from 161.0 to 154.2\,ms and the worst case rising from 305 to 348\,ms, at the cost of 27.4\% fewer completed learner updates, 1,182 against 1,627.
A 10\,Hz timer takes the latest image, the 100\,ms deadline runs from image
publication to CPU-ready action, and late actions hold the previous command.
Figure~\ref{fig:casestudy-synth} shows centered 30\,s means of all 6000 responses per arm, under 0.1\% of them invalid.

\noindent\textbf{Learning progress.} We measure learning with an offline imitation proxy return, since the car
lacks the LiDAR of the simulated pretraining scenarios. Following Ekya~\cite{bhardwaj2022ekya}, an accurate larger model labels the
recorded images offline. The proxy is the negative mean squared action error on $341$ held-out images  disjoint from the $1{,}024$ training samples.
In Figure~\ref{fig:casestudy-synth}, over ten snapshots per arm taken every 60\,s, \tool{} reaches a proxy
return close to the no-runtime arm's, ending slightly lower.

\section{Conclusion}

\tool{} makes the actor--learner coupling a tunable runtime parameter and
co-schedules both with a permit and slack-aware admission atop the
platform's scheduler; the refresh bound sets update lag with double buffering and refresh frequency on the copy path. On two Jetson boards, slack-aware admission is what turns GPU serialisation into timeliness: under the real refresh cost it keeps actor misses below $1\%$ on large networks while retaining most of the learner's throughput, and update splitting lets updates longer than the actor's slack make progress. A deployment sets the refresh bound from the control loop's miss budget on the copy path, or from the update lag it tolerates with double buffering.

\reinbibliography{references}

\begin{thebibliography}{28}
\providecommand{\natexlab}[1]{#1}
\providecommand{\url}[1]{\texttt{#1}}
\expandafter\ifx\csname urlstyle\endcsname\relax
  \providecommand{\doi}[1]{doi: #1}\else
  \providecommand{\doi}{doi: \begingroup \urlstyle{rm}\Url}\fi

\bibitem[Mnih et~al.(2015)Mnih, Kavukcuoglu, Silver, Rusu, Veness, Bellemare, Graves, Riedmiller, Fidjeland, Ostrovski, et~al.]{mnih2015human}
Volodymyr Mnih, Koray Kavukcuoglu, David Silver, Andrei~A Rusu, Joel Veness, Marc~G Bellemare, Alex Graves, Martin Riedmiller, Andreas~K Fidjeland, Georg Ostrovski, et~al.
\newblock Human-level control through deep reinforcement learning.
\newblock \emph{nature}, 518\penalty0 (7540):\penalty0 529--533, 2015.

\bibitem[Gog et~al.(2022)Gog, Kalra, Schafhalter, Gonzalez, and Stoica]{d3}
Ionel Gog, Sukrit Kalra, Peter Schafhalter, Joseph~E Gonzalez, and Ion Stoica.
\newblock D3: a dynamic deadline-driven approach for building autonomous vehicles.
\newblock In \emph{Proceedings of the Seventeenth European Conference on Computer Systems}, pages 453--471, 2022.

\bibitem[Kang et~al.(2021)Kang, Lee, Lee, Shin, and Chwa]{kang2021lalarand}
Woosung Kang, Kilho Lee, Jinkyu Lee, Insik Shin, and Hoon~Sung Chwa.
\newblock Lalarand: Flexible layer-by-layer cpu/gpu scheduling for real-time dnn tasks.
\newblock In \emph{2021 IEEE Real-Time Systems Symposium (RTSS)}, pages 329--341. IEEE, 2021.

\bibitem[Li et~al.(2021)Li, Zhang, Ding, Zhou, and Liu]{li2021efficient}
Zexin Li, Yuqun Zhang, Ao~Ding, Husheng Zhou, and Cong Liu.
\newblock Efficient algorithms for task mapping on heterogeneous cpu/gpu platforms for fast completion time.
\newblock \emph{Journal of Systems Architecture}, 114:\penalty0 101936, 2021.

\bibitem[Horgan et~al.(2018)Horgan, Quan, Budden, Barth-Maron, Hessel, van Hasselt, and Silver]{horgan2018distributed}
Dan Horgan, John Quan, David Budden, Gabriel Barth-Maron, Matteo Hessel, Hado van Hasselt, and David Silver.
\newblock Distributed prioritized experience replay.
\newblock In \emph{International Conference on Learning Representations}, 2018.
\newblock URL \url{https://openreview.net/forum?id=H1Dy---0Z}.

\bibitem[Espeholt et~al.(2018)Espeholt, Soyer, Munos, Simonyan, Mnih, Ward, Doron, Firoiu, Harley, Dunning, et~al.]{espeholt2018impala}
Lasse Espeholt, Hubert Soyer, Remi Munos, Karen Simonyan, Vlad Mnih, Tom Ward, Yotam Doron, Vlad Firoiu, Tim Harley, Iain Dunning, et~al.
\newblock Impala: Scalable distributed deep-rl with importance weighted actor-learner architectures.
\newblock In \emph{International conference on machine learning}, pages 1407--1416. PMLR, 2018.

\bibitem[Ramstedt and Pal(2019)]{ramstedt2019real}
Simon Ramstedt and Chris Pal.
\newblock Real-time reinforcement learning.
\newblock \emph{Advances in neural information processing systems}, 32, 2019.

\bibitem[Elliott et~al.(2013)Elliott, Ward, and Anderson]{elliott2013gpusync}
Glenn~A Elliott, Bryan~C Ward, and James~H Anderson.
\newblock Gpusync: A framework for real-time gpu management.
\newblock In \emph{2013 IEEE 34th Real-Time Systems Symposium}, pages 33--44. IEEE, 2013.

\bibitem[Zhou et~al.(2018)Zhou, Bateni, and Liu]{s3dnn}
Husheng Zhou, Soroush Bateni, and Cong Liu.
\newblock S\^{} 3dnn: Supervised streaming and scheduling for gpu-accelerated real-time dnn workloads.
\newblock In \emph{2018 IEEE Real-Time and Embedded Technology and Applications Symposium (RTAS)}, pages 190--201. IEEE, 2018.

\bibitem[Mnih et~al.(2016)Mnih, Badia, Mirza, Graves, Lillicrap, Harley, Silver, and Kavukcuoglu]{mnih2016asynchronous}
Volodymyr Mnih, Adria~Puigdomenech Badia, Mehdi Mirza, Alex Graves, Timothy Lillicrap, Tim Harley, David Silver, and Koray Kavukcuoglu.
\newblock Asynchronous methods for deep reinforcement learning.
\newblock In \emph{International conference on machine learning}, pages 1928--1937. PmLR, 2016.

\bibitem[Wang et~al.(2024)Wang, Liu, Wong, and Kim]{gcaps_ecrts24}
Yidi Wang, Cong Liu, Daniel Wong, and Hyoseung Kim.
\newblock {GCAPS: GPU Context-Aware Preemptive Priority-Based Scheduling for Real-Time Tasks}.
\newblock In Rodolfo Pellizzoni, editor, \emph{36th Euromicro Conference on Real-Time Systems (ECRTS 2024)}, volume 298 of \emph{Leibniz International Proceedings in Informatics (LIPIcs)}, pages 14:1--14:25, Dagstuhl, Germany, 2024. Schloss Dagstuhl -- Leibniz-Zentrum f{\"u}r Informatik.
\newblock ISBN 978-3-95977-324-9.
\newblock \doi{10.4230/LIPIcs.ECRTS.2024.14}.
\newblock URL \url{https://drops.dagstuhl.de/entities/document/10.4230/LIPIcs.ECRTS.2024.14}.

\bibitem[Han et~al.(2022)Han, Zhang, Chen, and Chen]{han2022reef}
Mingcong Han, Hanze Zhang, Rong Chen, and Haibo Chen.
\newblock Microsecond-scale preemption for concurrent {GPU-accelerated} {DNN} inferences.
\newblock In \emph{16th USENIX Symposium on Operating Systems Design and Implementation (OSDI 22)}, pages 539--558, Carlsbad, CA, July 2022. USENIX Association.
\newblock ISBN 978-1-939133-28-1.
\newblock URL \url{https://www.usenix.org/conference/osdi22/presentation/han}.

\bibitem[Paull et~al.(2017)Paull, Tani, Ahn, Alonso-Mora, Carlone, Cap, Chen, Choi, Dusek, Fang, Hoehener, Liu, Novitzky, Okuyama, Pazis, Rosman, Varricchio, Wang, Yershov, Zhao, Benjamin, Carr, Zuber, Karaman, Frazzoli, Del~Vecchio, Rus, How, Leonard, and Censi]{paull2017duckietown}
Liam Paull, Jacopo Tani, Heejin Ahn, Javier Alonso-Mora, Luca Carlone, Michal Cap, Yu~Fan Chen, Changhyun Choi, Jeff Dusek, Yajun Fang, Daniel Hoehener, Shih-Yuan Liu, Michael Novitzky, Igor~Franzoni Okuyama, Jason Pazis, Guy Rosman, Valerio Varricchio, Hsueh-Cheng Wang, Dmitry Yershov, Hang Zhao, Michael Benjamin, Christopher Carr, Maria Zuber, Sertac Karaman, Emilio Frazzoli, Domitilla Del~Vecchio, Daniela Rus, Jonathan How, John Leonard, and Andrea Censi.
\newblock Duckietown: An open, inexpensive and flexible platform for autonomy education and research.
\newblock In \emph{2017 IEEE International Conference on Robotics and Automation (ICRA)}, pages 1497--1504. IEEE, May 2017.
\newblock \doi{10.1109/icra.2017.7989179}.
\newblock URL \url{http://dx.doi.org/10.1109/ICRA.2017.7989179}.

\bibitem[Van~Hasselt et~al.(2016)Van~Hasselt, Guez, and Silver]{van2016deep}
Hado Van~Hasselt, Arthur Guez, and David Silver.
\newblock Deep reinforcement learning with double q-learning.
\newblock \emph{Proceedings of the AAAI Conference on Artificial Intelligence}, 30\penalty0 (1), March 2016.
\newblock ISSN 2159-5399.
\newblock \doi{10.1609/aaai.v30i1.10295}.
\newblock URL \url{http://dx.doi.org/10.1609/aaai.v30i1.10295}.

\bibitem[Bellemare et~al.(2017)Bellemare, Dabney, and Munos]{bellemare2017distributional}
Marc~G. Bellemare, Will Dabney, and R{\'e}mi Munos.
\newblock A distributional perspective on reinforcement learning.
\newblock In Doina Precup and Yee~Whye Teh, editors, \emph{Proceedings of the 34th International Conference on Machine Learning}, volume~70 of \emph{Proceedings of Machine Learning Research}, pages 449--458. PMLR, 06--11 Aug 2017.
\newblock URL \url{https://proceedings.mlr.press/v70/bellemare17a.html}.

\bibitem[Haarnoja et~al.(2018)Haarnoja, Zhou, Abbeel, and Levine]{haarnoja2018soft}
Tuomas Haarnoja, Aurick Zhou, Pieter Abbeel, and Sergey Levine.
\newblock Soft actor-critic: Off-policy maximum entropy deep reinforcement learning with a stochastic actor.
\newblock In Jennifer Dy and Andreas Krause, editors, \emph{Proceedings of the 35th International Conference on Machine Learning}, volume~80 of \emph{Proceedings of Machine Learning Research}, pages 1861--1870. PMLR, 10--15 Jul 2018.
\newblock URL \url{https://proceedings.mlr.press/v80/haarnoja18b.html}.

\bibitem[Fujimoto et~al.(2018)Fujimoto, van Hoof, and Meger]{fujimoto2018addressing}
Scott Fujimoto, Herke van Hoof, and David Meger.
\newblock Addressing function approximation error in actor-critic methods.
\newblock In Jennifer Dy and Andreas Krause, editors, \emph{Proceedings of the 35th International Conference on Machine Learning}, volume~80 of \emph{Proceedings of Machine Learning Research}, pages 1587--1596. PMLR, 10--15 Jul 2018.
\newblock URL \url{https://proceedings.mlr.press/v80/fujimoto18a.html}.

\bibitem[Gu et~al.(2017)Gu, Holly, Lillicrap, and Levine]{gu2017deep}
Shixiang Gu, Ethan Holly, Timothy Lillicrap, and Sergey Levine.
\newblock Deep reinforcement learning for robotic manipulation with asynchronous off-policy updates.
\newblock In \emph{2017 IEEE International Conference on Robotics and Automation (ICRA)}, pages 3389--3396. IEEE, May 2017.
\newblock \doi{10.1109/icra.2017.7989385}.
\newblock URL \url{http://dx.doi.org/10.1109/ICRA.2017.7989385}.

\bibitem[Chen et~al.(2025)Chen, Chen, Li, Jiang, Wan, He, Ran, Gu, Li, Xie, and Ray]{chen2025benchmarking}
Simin Chen, Yiming Chen, Zexin Li, Yifan Jiang, Zhongwei Wan, Yixin He, Dezhi Ran, Tianle Gu, Haizhou Li, Tao Xie, and Baishakhi Ray.
\newblock Benchmarking large language models under data contamination: A survey from static to dynamic evaluation.
\newblock In Christos Christodoulopoulos, Tanmoy Chakraborty, Carolyn Rose, and Violet Peng, editors, \emph{Proceedings of the 2025 Conference on Empirical Methods in Natural Language Processing}, pages 10080--10098, Suzhou, China, November 2025. Association for Computational Linguistics.
\newblock ISBN 979-8-89176-332-6.
\newblock \doi{10.18653/v1/2025.emnlp-main.511}.
\newblock URL \url{https://aclanthology.org/2025.emnlp-main.511/}.

\bibitem[Choi et~al.(2021)Choi, Kim, and Zhu]{choi2021toward}
Hyunjong Choi, Hyoseung Kim, and Qi~Zhu.
\newblock Toward practical weakly hard real-time systems: A job-class-level scheduling approach.
\newblock \emph{IEEE Internet of Things Journal}, 8\penalty0 (8):\penalty0 6692--6708, April 2021.
\newblock ISSN 2372-2541.
\newblock \doi{10.1109/jiot.2021.3058215}.
\newblock URL \url{http://dx.doi.org/10.1109/JIOT.2021.3058215}.

\bibitem[Xiang and Kim(2019)]{bib:RTSS19:Xiang}
Yecheng Xiang and Hyoseung Kim.
\newblock Pipelined data-parallel cpu/gpu scheduling for multi-dnn real-time inference.
\newblock In \emph{2019 IEEE Real-Time Systems Symposium (RTSS)}, pages 392--405. IEEE, December 2019.
\newblock \doi{10.1109/rtss46320.2019.00042}.
\newblock URL \url{http://dx.doi.org/10.1109/RTSS46320.2019.00042}.

\bibitem[Bateni and Liu(2018)]{bateni2018apnet}
Soroush Bateni and Cong Liu.
\newblock Apnet: Approximation-aware real-time neural network.
\newblock In \emph{2018 IEEE Real-Time Systems Symposium (RTSS)}, pages 67--79. IEEE, December 2018.
\newblock \doi{10.1109/rtss.2018.00017}.
\newblock URL \url{http://dx.doi.org/10.1109/RTSS.2018.00017}.

\bibitem[Li et~al.(2023)Li, Samanta, Li, Soltoggio, Kim, and Liu]{li2023r3}
Zexin Li, Aritra Samanta, Yufei Li, Andrea Soltoggio, Hyoseung Kim, and Cong Liu.
\newblock $\mathrm{R}^{3}$: On-device real-time deep reinforcement learning for autonomous robotics.
\newblock In \emph{2023 IEEE Real-Time Systems Symposium (RTSS)}, pages 131--144. IEEE, December 2023.
\newblock \doi{10.1109/rtss59052.2023.00021}.
\newblock URL \url{http://dx.doi.org/10.1109/RTSS59052.2023.00021}.

\bibitem[Ho et~al.(2013)Ho, Cipar, Cui, Lee, Kim, Gibbons, Gibson, Ganger, and Xing]{ho2013more}
Qirong Ho, James Cipar, Henggang Cui, Seunghak Lee, Jin~Kyu Kim, Phillip~B. Gibbons, Garth~A Gibson, Greg Ganger, and Eric Xing.
\newblock More effective distributed ml via a stale synchronous parallel parameter server.
\newblock In C.J. Burges, L.~Bottou, M.~Welling, Z.~Ghahramani, and K.~Weinberger, editors, \emph{Advances in Neural Information Processing Systems}, volume~26. Curran Associates, Inc., 2013.
\newblock URL \url{https://proceedings.neurips.cc/paper_files/paper/2013/file/b7bb35b9c6ca2aee2df08cf09d7016c2-Paper.pdf}.

\bibitem[Liang et~al.(2018)Liang, Liaw, Nishihara, Moritz, Fox, Goldberg, Gonzalez, Jordan, and Stoica]{liang2018rllib}
Eric Liang, Richard Liaw, Robert Nishihara, Philipp Moritz, Roy Fox, Ken Goldberg, Joseph Gonzalez, Michael Jordan, and Ion Stoica.
\newblock {RL}lib: Abstractions for distributed reinforcement learning.
\newblock In Jennifer Dy and Andreas Krause, editors, \emph{Proceedings of the 35th International Conference on Machine Learning}, volume~80 of \emph{Proceedings of Machine Learning Research}, pages 3053--3062. PMLR, 10--15 Jul 2018.
\newblock URL \url{https://proceedings.mlr.press/v80/liang18b.html}.

\bibitem[Petrenko et~al.(2020)Petrenko, Huang, Kumar, Sukhatme, and Koltun]{petrenko2020sample}
Aleksei Petrenko, Zhehui Huang, Tushar Kumar, Gaurav Sukhatme, and Vladlen Koltun.
\newblock Sample factory: Egocentric 3{D} control from pixels at 100000 {FPS} with asynchronous reinforcement learning.
\newblock In Hal~Daumé III and Aarti Singh, editors, \emph{Proceedings of the 37th International Conference on Machine Learning}, volume 119 of \emph{Proceedings of Machine Learning Research}, pages 7652--7662. PMLR, 13--18 Jul 2020.
\newblock URL \url{https://proceedings.mlr.press/v119/petrenko20a.html}.

\bibitem[Schulman et~al.(2017)Schulman, Wolski, Dhariwal, Radford, and Klimov]{schulman2017proximal}
John Schulman, Filip Wolski, Prafulla Dhariwal, Alec Radford, and Oleg Klimov.
\newblock Proximal policy optimization algorithms, 2017.
\newblock URL \url{https://arxiv.org/abs/1707.06347}.

\bibitem[Bhardwaj et~al.(2022)Bhardwaj, Xia, Ananthanarayanan, Jiang, Shu, Karianakis, Hsieh, Bahl, and Stoica]{bhardwaj2022ekya}
Romil Bhardwaj, Zhengxu Xia, Ganesh Ananthanarayanan, Junchen Jiang, Yuanchao Shu, Nikolaos Karianakis, Kevin Hsieh, Paramvir Bahl, and Ion Stoica.
\newblock Ekya: Continuous learning of video analytics models on edge compute servers.
\newblock In \emph{19th USENIX Symposium on Networked Systems Design and Implementation (NSDI 22)}, pages 119--135, Renton, WA, April 2022. USENIX Association.
\newblock ISBN 978-1-939133-27-4.
\newblock URL \url{https://www.usenix.org/conference/nsdi22/presentation/bhardwaj}.

\end{thebibliography}
\end{document}